\documentclass[runningheads]{llncs}

\usepackage{graphicx}
\usepackage{booktabs}

\usepackage[accsupp]{axessibility}  

\usepackage[pagebackref,breaklinks,colorlinks,citecolor=blue]{hyperref}

\usepackage{orcidlink}

\begin{document}

\title{CSS-Segment: 2nd Place Report of LSVOS Challenge VOS Track} 

\author{Jinming Chai\inst{1} \and
Qin Ma\inst{1} \and
Junpei Zhang\inst{1} \and
Licheng Jiao\inst{1} \and
Fang Liu\inst{1}}

\authorrunning{J. Chai et al.}

\institute{Intelligent Perception and Image Understanding Lab, Xidian University\\
Team: yuanjie}

\maketitle

\begin{abstract}
Video object segmentation is a challenging task that serves as the cornerstone of numerous downstream applications, including video editing\cite{video editing} and autonomous driving. In this technical report, we briefly introduce the solution of our team "yuanjie" for video object segmentation in the 6-th LSVOS Challenge VOS Track at ECCV 2024. We believe that our proposed CSS-Segment will perform better in videos of complex object motion and long-term presentation. In this report, we successfully validated the effectiveness of the CSS-Segment in video object segmentation. Finally, our method achieved a J\&F score of 80.84 in and test phases, and ultimately ranked 2nd in the 6-th LSVOS Challenge VOS Track at ECCV 2024.
\keywords{VOS \and Finetune \and Inference \and Muti Level Fusion }
\end{abstract}

\section{Introduction}
\label{sec:intro}
Video Object Segmentation (VOS) is a complex task derived from image object segmentation, requiring the classification and segmentation of all objects in a video. With its broad applicability in various downstream tasks, including video understanding, editing, and autonomous driving, VOS has garnered significant attention in recent years. This year's LSVOS Challenge includes two tracks: Referring Video Object Segmentation (RVOS) and Video Object Segmentation (VOS). The RVOS track has been upgraded with the newly introduced MeViS\cite{MeViS} dataset, which offers more challenging motion-guided language expressions and more complex videos compared to the previous Refer-Youtube-VOS dataset. Similarly, the VOS track now uses the MOSE\cite{MOSE} dataset instead of Youtube-VOS, featuring scenarios where objects disappear and reappear, small and difficult-to-detect objects, heavy occlusions, and more, making this year's competition significantly more complex than before.

Video Object Segmentation (VOS) extends the challenges of image segmentation to the spatio-temporal domain, where entities must be accurately classified and segmented across video frames. Unlike images, video entities undergo significant changes in appearance due to motion, deformation, occlusion, and lighting variations, compounded by the lower quality often seen in videos due to factors like camera motion, blur, and resolution constraints. Efficiently processing these large frame sequences further complicates the task.

Recent VOS approaches have embraced a memory-based paradigm, where a memory representation, computed from past segmented frames, aids in segmenting new query frames by "reading" from this memory. Most approaches rely on pixel-level matching for this memory reading, mapping each query pixel to a combination of memory pixels. While effective in simpler scenarios, pixel-level matching can suffer from noise and inconsistencies, especially in the presence of occlusions and distractors, leading to reduced performance in more complex environments. For instance, evaluations on the MOSE\cite{MOSE} dataset show that recent VOS methods perform over 20 points lower in J\&F compared to simpler benchmarks like DAVIS\cite{DAVIS}. To address these challenges, there is a growing interest in developing more robust techniques, such as object-level memory reading, which aim to improve segmentation consistency and reduce susceptibility to distractors in complex video scenes.

Given that LVOS\cite{LVOS} focuses on long-term videos with complex object motion and long-term reappearance, conventional models struggle to perform well on such data. To address this challenge, models must possess robust capabilities that allow for better image representation while also being able to store longer-term memories to handle these complex scenarios effectively. In the VOS task, the Cutie\cite{Cutie}, SAM, and SAM2\cite{SAM2} models have all demonstrated outstanding performance, each excelling in different aspects of network architecture. Therefore, in this technical report, we propose the CSS-Segment by efficiently integrating the advantageous modules of Cutie, SAM, and SAM2. Validation on the challenge's test set shows that CSS-Segment delivers exceptional performance.

\section{Method}
\begin{figure}[h]
\centering
\includegraphics[width=0.8\textwidth]{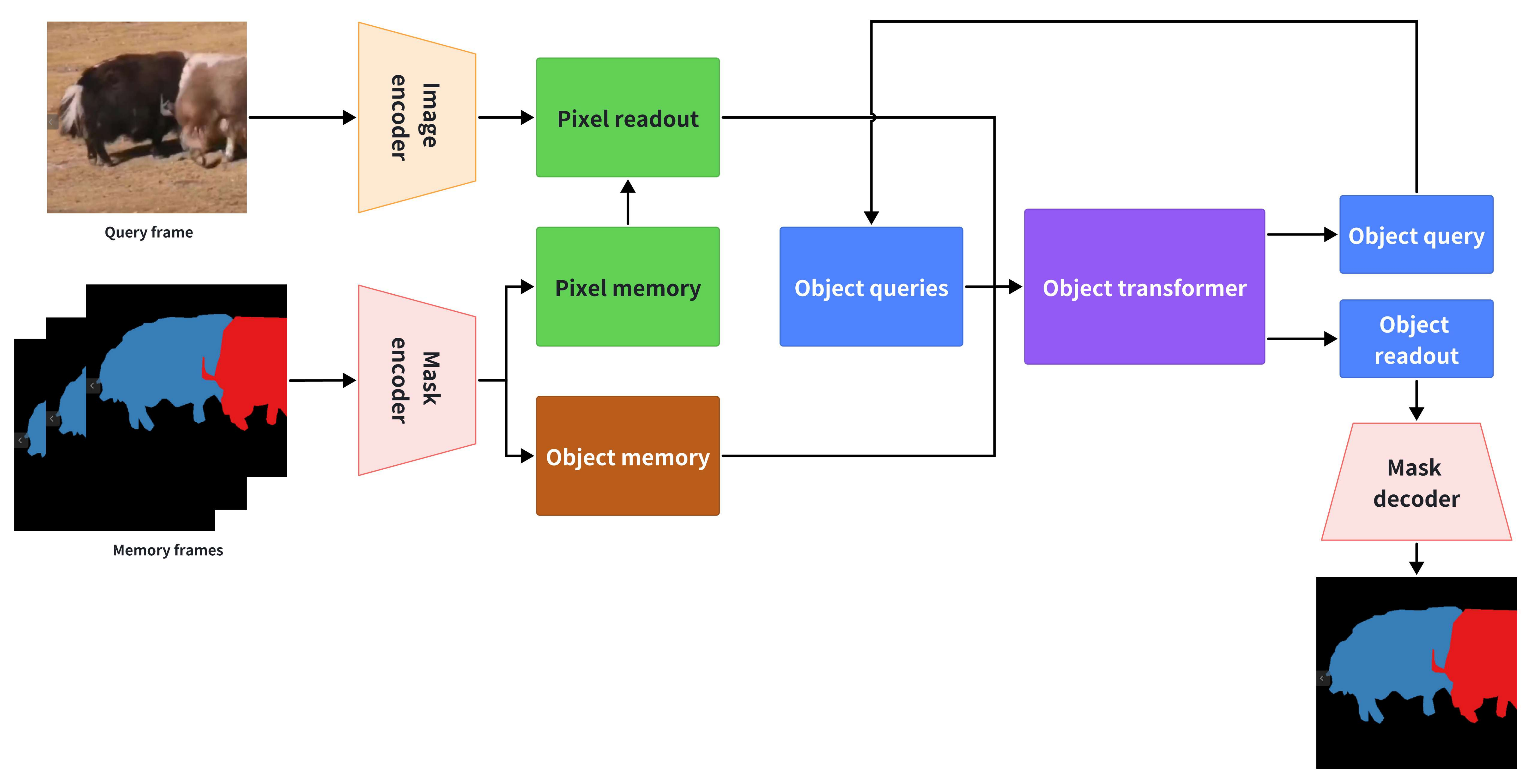} 
\caption{Workflow of the CSS-Segment. Image encoder is a streaming approach, consuming video frames as they become available. Mask encoder using convolutions and
summed element-wise with the image embedding. We store pixel memory and object memory representations from past segmented (memory) frames. Pixel memory is retrieved for the query frame as pixel readout, which bidirectionally interacts with object queries and object memory in the object transformer. The object transformer blocks enrich the pixel feature with object-level semantics and produce the final object readout for decoding into the output mask.} 
\label{fig:1} 
\end{figure}

The proposed restoration framework contains four main steps, as shown in Fig\ref{fig:1}.

\begin{enumerate}
\setlength{\itemindent}{2em}
\setlength{\leftmargin}{2em}
\item Image Encoder
\item Mask Encoder
\item Object Transformer
\item Object Memory
\end{enumerate}

\subsection{Image Encoder}
The image encoder used in our framework is inspired by the design principles of SAM2 and is tailored for real-time processing of arbitrarily long videos\cite{long videos}. Unlike the ResNet50-based encoder used in the Cutie model, which may struggle with long sequences, our image encoder leverages a streaming approach. It processes video frames as they become available and is executed only once for the entire interaction. This design allows the encoder to provide unconditioned tokens (feature embeddings) that represent each frame effectively.

We utilize a hierarchical MAE (He et al., 2022) Hiera (Ryali et al., 2023; Bolya et al., 2023) image encoder, which is specifically designed to handle multiscale features. This hierarchical structure enhances the encoder's ability to capture and represent complex, long-term video sequences more effectively than traditional models. By integrating multiscale features during decoding, our approach achieves superior data representation for long-sequence motion videos, addressing the limitations observed with the ResNet50-based encoder in the Cutie model.

\subsection{Mask Encoder}
The design of our mask encoder is primarily inspired by the mask encoder used in SAM, offering a notable advancement over the ResNet18-based mask encoder utilized in the Cutie model. Our mask encoder integrates dense prompts (i.e., masks) with the image embeddings through a series of sophisticated convolutional operations. Specifically, masks are initially processed at a resolution 4× lower than the input image, followed by further downscaling using two 2×2 convolutions with stride 2, featuring output channels of 4 and 16 respectively. This is complemented by a final 1×1 convolution that adjusts the channel dimension to 256. The entire process is enhanced with GELU activations and layer normalization at each stage.

This approach allows for a more precise spatial\cite{spatial} correspondence between the masks and image embeddings, ensuring that the mask information is effectively incorporated into the image features. Unlike the ResNet18-based mask encoder in the Cutie model, which relies on standard convolutional layers without such detailed spatial management, our design ensures more accurate and nuanced feature representation. The addition of a learned embedding to represent "no mask" scenarios further enhances the encoder's capability to handle diverse and complex input conditions. This results in improved data representation and segmentation performance, demonstrating the superiority of our mask encoder in capturing and integrating detailed mask information compared to the Cutie model's approach.

\subsection{Object Transformer}

The Object Transformer, processes an initial readout $R_0 \in \mathbb{R}^{H \times W \times C}$, a set of $N$ end-to-end trained object queries $X \in \mathbb{R}^{N \times C}$, and object memory $S \in \mathbb{R}^{N \times C}$. It integrates these with $L$ transformer blocks to produce the final output. Here, $H$ and $W$ denote the image dimensions after encoding with a stride of 16. 

Before the first transformer block, the static object queries are summed with the dynamic object memory: $X_0 = X + S$. Each transformer block allows the object queries $X_{l-1}$ to attend to the readout $R_{l-1}$ bidirectionally, and vice versa, updating the queries to $X_l$ and the readout to $R_l$. The final readout $R_L$ of the last block is the output of the Object Transformer.

\section{Object Memory}

The object memory, denoted as $S \in \mathbb{R}^{N \times C}$, stores a compact set of $N$ vectors that provide a high-level summary of the target object. This object memory is utilized in the Object Transformer (see Section~\ref{sec:object_transformer}) to offer target-specific features.

To compute $S$, we perform mask-pooling over all encoded object features. Specifically, given the object features $U \in \mathbb{R}^{THW \times C}$ and $N$ pooling masks $\{W_q \in [0, 1]^{THW} \mid 0 < q \leq N\}$, each mask $W_q$ is used to aggregate the features in $U$ into a summary vector for the object memory.

This pooling process ensures that the object memory captures relevant information from the encoded features, which is then leveraged for effective object representation in the transformer.

\section{Experiment}

We fine-tune the CSS-Segment network using the MOSE and LVOS datasets, where the Image Encoder, Mask Encoder, and other modules of CSS-Segment are initialized with pre-trained weights from SAM2, SAM, and Cutie, respectively. During inference, data augmentation techniques are employed, and the final results are subjected to pixel-level and video-level fusion.

\subsection{Dataset}

In our experiments, we utilize the MOSE and LVOS datasets to fine-tune segmentation model. The MOSE dataset provides high-precision annotations for multi-object segmentation, focusing on diverse scenes with overlapping objects and varying sizes. The LVOS dataset offers comprehensive video sequences with detailed object annotations, supporting the segmentation and tracking\cite{tracking} of objects across frames. These datasets enable us to rigorously assess the performance of our models in handling both static and dynamic object segmentation tasks.

\subsection{Fine-tune}

Due to the large volume of data in the MOSE and LVOS datasets, training from scratch is both time-consuming and challenging. Therefore, we chose to fine-tune the CSS-Segment network using open-source pre-trained weights. Specifically, the Image Encoder, Mask Encoder, and other modules of the network are initialized with pre-trained weights from SAM2, SAM, and Cutie, respectively. The fine-tuning process is performed using four NVIDIA GeForce RTX 3090 GPUs.

\subsection{Inference}
During model inference, we introduce three key innovations. First, we adjust the hyperparameters to accommodate the varying frame lengths of different videos in the test set. Specifically, we divide the 150 videos into two groups based on frame count: for videos with fewer than 200 frames, we set the parameters to \texttt{max\_mem\_frames=15}, \texttt{min\_mem\_frames=14}, and \texttt{top\_k=30}; for videos with more than 200 frames, we set them to \texttt{max\_mem\_frames=45}, \texttt{min\_mem\_frames=40}, and \texttt{top\_k=40}. Second, we employ multi-scale fusion by resizing video frames to 480, 660, 800, and 1000 pixels during inference, and then fuse the predictions across these scales to improve accuracy. Third, we apply horizontal flipping to the videos during inference to enhance the model's accuracy by increasing variability and improving generalization.

\subsection{Muti level Fusion}
In the result fusion process, we perform both pixel-level and video-level fusion. Pixel-level fusion involves aggregating all inference results by applying weighted voting at each pixel. This approach combines predictions from different models or runs to produce a more accurate result for each pixel. Video-level fusion is achieved by analyzing the log files obtained from CodaLab submissions. Specifically, we select the video with the highest J\&F score for each of the 150 videos in the test set from multiple logs. This method ensures that the most accurate video predictions are chosen for the final evaluation.

\section{Conclusion}
In this technical report, we introduce CSS-Segment, designed for the 6th LSVOS Challenge VOS Track at ECCV 2024. CSS-Segment integrates modules from Cutie, SAM, and SAM2 to advance video object segmentation. The system employs a streaming approach for the Image Encoder and applies convolutional operations and element-wise summation with image embeddings for the Mask Encoder. Pixel and object memory representations from previous frames are utilized, with pixel memory providing readout for query frames that interact bidirectionally with object queries and memory in the object transformer, enriching features with object-level semantics and generating the final readout for mask decoding. Fine-tuning was performed on the MOSE and LVOS datasets, using pre-trained weights from SAM2, SAM, and Cutie. Data augmentation and both pixel-level and video-level fusion techniques were applied during inference. Our method achieved a J\&F score of 80.84 and ranked 2nd in the challenge.

%
%

\begin{thebibliography}{99}

\bibitem{MeViS}
Ding, H., Liu, C., He, S., Jiang, X., \& Loy, C. C. (2023). MeViS: A large-scale benchmark for video segmentation with motion expressions. In Proceedings of the IEEE/CVF International Conference on Computer Vision (pp. 2694-2703).

\bibitem{MOSE}
Ding, H., Liu, C., He, S., Jiang, X., Torr, P. H., \& Bai, S. (2023). MOSE: A new dataset for video object segmentation in complex scenes. In Proceedings of the IEEE/CVF International Conference on Computer Vision (pp. 20224-20234).

\bibitem{LVOS}
Hong, L., Liu, Z., Chen, W., Tan, C., Feng, Y., Zhou, X., Guo, P., Li, J., Chen, Z., Gao, S., et al.: LVOS: A Benchmark for Large-scale Long-term Video Object Segmentation. arXiv preprint arXiv:2404.19326 (2024)

\bibitem{DAVIS}
Perazzi, F., Liu, S., Pont-Tuset, J., Ham, D., Arbeláez, P., Rother, C.: Davis: A Dataset for Video Object Segmentation. In: CVPR (2016)

\bibitem{SAM2}
Lin, T.Y., Goyal, P., Girshick, R., He, K., Dollár, P.: SAM2: Segment Anything Model 2. In: CVPR (2024)

\bibitem{Cutie}
Zhou, X., Wang, H., Yang, Y., Zhang, J.: Cutie: A Comprehensive Toolkit for Efficient Video Object Segmentation. In: ECCV (2022)

\bibitem{tracking}
Xie, S., Zheng, X., Li, Y., Han, Y.: Tracking Objects in Long Videos with Temporal Consistency. In: ECCV (2022)

\bibitem{long videos}
Chen, X., Zhang, Z., Lu, Z., Yang, M.: Long-term Video Object Segmentation with Memory Augmentation. In: CVPR (2023)

\bibitem{spatial}
Li, Z., Wang, X., Liu, L.: Spatiotemporal Fusion for Video Object Segmentation. In: CVPR (2021)

\bibitem{video editing}
Kim, J., Jeong, D., Park, S., Lee, S.: Video Editing Based on Video Object Segmentation and Temporal Consistency. In: CVPR (2024)


\end{thebibliography}


\end{document}